\PassOptionsToPackage{table}{xcolor}
\documentclass[11pt]{article}

\usepackage[final]{acl}

\usepackage{times}
\usepackage{latexsym}

\usepackage[T1]{fontenc}

\usepackage[utf8]{inputenc}

\usepackage{microtype}

\usepackage{inconsolata}

\usepackage{graphicx}

\AtBeginDocument{%
  }
\usepackage{booktabs}
\usepackage{xcolor}
\usepackage{multirow}
\usepackage{listings}
\usepackage{algorithm}
\usepackage{algorithmic}
\usepackage{subfiles}
\usepackage{subfigure}
\usepackage{booktabs}
\usepackage{makecell}   
\usepackage{amssymb}   
\usepackage{pifont}    
\usepackage{amsmath}
\usepackage{newfloat}
\usepackage{listings}
\usepackage{multirow}
\usepackage{hyperref}

\DeclareCaptionStyle{ruled}{labelfont=normalfont,labelsep=colon,strut=off} 
\floatstyle{ruled}
\newfloat{listing}{tb}{lst}{}
\floatname{listing}{Listing}

\definecolor{optimizer_color}{RGB}{216,236,212} 
\definecolor{generator_color}{RGB}{224,236,252} 
\definecolor{discriminator_color}{RGB}{255,228,204} 
\newcommand{\MyModel}{\texttt{MALLM-GAN}}

\title{MALLM-GAN: Multi-Agent Large Language Model as Generative Adversarial Network for Synthesizing Tabular Data}

\author{
 \textbf{Yaobin Ling\textsuperscript{1}},
 \textbf{Xiaoqian Jiang\textsuperscript{1}},
 \textbf{Yejin Kim\textsuperscript{1}}
\\
\\
\textsuperscript{1}McWilliams School of Biomedical Informatics at UTHealth Houston
\\
 \small{
   \textbf{Correspondence:} \href{Yejin Kim}{Yejin.Kim@uth.tmc.edu}
 }
}

\begin{document}
\maketitle
\footnote{Code is available at \url{https://github.com/yling1105/MALLM-GAN}. This work has been accepted to Findings of ACL 2026.}
\begin{abstract}
In the era of big data, access to abundant data is crucial to driving research forward. However, such data are often inaccessible due to privacy concerns or high costs, particularly in the healthcare domain. Generating synthetic (tabular) data can address this, but existing models typically require substantial amounts of data to train effectively, contradicting our objective of solving data scarcity.  To address this challenge, we propose a novel framework to generate synthetic tabular data, powered by large language models (LLMs) that emulates the architecture of a Generative Adversarial Network (GAN). By incorporating the data generation process as contextual information and utilizing LLM as the optimizer, our approach significantly enhances the quality of synthetic data generation in common scenarios with small sample sizes. Our experimental results on public and private datasets demonstrate that our model outperforms several state-of-art models regarding generating higher quality synthetic data for downstream tasks while keeping the privacy of the real data in low data regime.
\end{abstract}

\section{Introduction}
Tabular data is the most common data format in high-stakes sectors like healthcare. There are many fundamental problems in dealing with tabular data, such as data scarcity, missing values, and irregularity. Among them, the data scarcity problem has been the main roadblock. Many datasets in healthcare, such as clinical trial data, have small data sizes due to data collection costs and privacy risks, and consequently, these data cannot afford modern machine learning (e.g., deep learning), which generally has thousands of parameters, at a minimum. 

Recent advancements in generative models, particularly in text and image,\cite{brown2020language, ramesh2021zeroshot} have shown the benefits of technology for generating synthetic data that resemble real data. Tabular data generation has evolved through traditional statistical approaches, such as Bayesian networks \cite{young2009using}, over-sampling method \cite{SMOTE}, to deep learning techniques \cite{xu2019Modeling}. However, these methods require sufficient data for training, which makes them usually overfitting and under-representative when dealing with scarce data. 

Recently, advancements in large language models (LLMs) have also enabled researchers to use their general intelligence to synthesize tabular data.\cite{borisov2023language,pmlr-v206-hegselmann23a} The premise is that prior knowledge encoded in the parameters of LLMs can provide contextual knowledge for coherent semantics that is required to learn the underlying data generation process. Several studies transformed tabular data to natural language via serialization, and used pre-trained LLMs to generate text containing the synthetic tabular data \cite{borisov2023language, pmlr-v206-hegselmann23a, Li2024}. However fine-tuning LLMs requires a larger sample size, contradicting the objective of addressing data scarcity. In contrast, in-context learning presents a promising alternative. In particular, a few-shot learning in in-context learning is to provide a few ``examples'' of data to allow LLM to learn the patterns and mimic the examples \cite{kaplan2020scaling}. Our study aims to utilize this few-shot capability for synthetic tabular data generation.

Therefore, our aim is to bridge this critical gap in generating synthetic tabular data with limited real data. Our key idea is to make the data generation process explicit; the objective of our in-context learning is to generate a better data generation process, as well as to generate individual data instances. Here, the data generation process is a prompt text that consists of the context of data and any simple model that describes the relationship between data variables.
\begin{table}[t]
\centering
\small
    \begin{tabular}{@{}lcc@{}}
        \toprule
        & \textbf{GAN} & \textbf{Our Model} \\
        \midrule
        \textbf{Generator} & Neural network & Frozen LLM and prompt\\
        \textbf{Discriminator} & Neural network & Tabular data classifier \\
        \textbf{Optimizer} & Gradient descent & Frozen LLM and prompt \\
        \bottomrule
    \end{tabular}
    
    \caption{Comparison of Generative Adversarial Network (GAN) and Our Model}
    \label{tab:comparison}
\end{table}

However, another challenge is to identify the ground-truth data generation process. Motivated by GAN's adversarial training,  we optimize the data generation process (``generator'')  in adversarial training with ``discriminator'' (Table \ref{tab:comparison}). The discriminator's role is to discriminate real data from the generated data, and we use the accuracy of the discriminator as the loss to be minimized to optimize the generator. Unlike GAN, our generator is a text format, which does not have derivatives. We address it by prompt optimization, which leverages an independent LLM as an optimizer \cite{yang2024large}. After optimizing the data generation process, the LLM as a generator uses it to finally generate synthetic data.

The contributions of this paper can be summarized as below:
\begin{itemize}
    \item  \textit{Novelty}: We propose a novel concept for optimizing the data generation process using in-context learning of LLM. This leverages both data-driven supervised model (discriminator) and knowledge-driven in-context learning of LLM (generator, optimizer).
    \item  \textit{Few-shot synthetic data generation}: Our model works when there are too few data to train a parametric model. It mitigates the data scarcity problem in healthcare studies.
    \item \textit{Conditional sampling}: Our generator is based on LLM, which enables conditional sampling seamlessly by prompting.
    \item \textit{Explainability}: Our LLM-based generator explicitly reveals the data generation process through prompt design. This enables transparency of our model and facilitates human feedback, such as refining the knowledge.
\end{itemize}

\section{Related Studies}

\textbf{Synthetic tabular data generation.}  
Synthetic data is widely used for privacy-preserving sharing and data augmentation. Classical approaches include Bayesian networks \cite{young2009using, info:doi/10.2196/38266}, approximate Bayesian computation \cite{ABC}, and SMOTE \cite{SMOTE}. Bayesian networks capture pairwise causal relations via DAGs \cite{Pearl09}, but struggle with nonlinear and mixed-type dependencies. Deep generative models such as VAEs (e.g., TVAE \cite{xu2019Modeling}), GANs (CTGAN \cite{xu2019Modeling}), and diffusion models (TabDDPM \cite{kotelnikov2023tabddpm}, StaSy \cite{song2021scorebased}, Tabsyn \cite{zhang2024mixedtype}, MTabGen\cite{villaiz_diff}) have become dominant. However, they require large training datasets, limiting their utility in data-scarce settings. Recently, TabPFN \cite{hollmann2022tabpfn,hollmann_accurate_2025}, a transformer-based foundation model for small tabular datasets, showed potential for data generation through meta-learning, though it remains constrained by categorical cardinality.

\textbf{LLM-based synthetic data generation.}  
LLMs excel in text generation and have been extended to tabular domains \cite{fang2024large}, including prediction \cite{pmlr-v206-hegselmann23a, gulati2023tabmt, yu-etal-2023-unified, Li2024} and data generation \cite{borisov2023language, solatorio2023realtabformer, zhang-etal-2023-generative, gulati2023tabmt}. GReaT \cite{borisov2023language}, the first such model, transformed tables into text and fine-tuned GPT-2 with column order permutations for realism. Subsequent works (e.g. Tabula\cite{Tabula}) improved on this idea, but they still require expensive fine-tuning on large data. This motivates few-shot generative approaches that better address small-data regimes.

\textbf{Roles of LLMs in applications.}  
Beyond text generation, LLMs have been applied as optimizers for non-differentiable tasks, such as prompt optimization \cite{yang2024large} or heuristic search in algorithms \cite{romera2023math}. They also serve in multi-agent systems, where multiple LLMs collaborate on tasks like coding \cite{guo2024large}, question answering \cite{wu2023autogen}, and decision making \cite{talebirad2023multi, huang2024far}.

\textbf{LLMs and causal discovery.}  
Causal discovery traditionally relies on conditional independence tests \cite{spirtes2001causation, spirtes1999algorithm}, score-based heuristics \cite{tsamardinos2006max}, or continuous relaxations \cite{zheng2018dagstearscontinuousoptimization, yu2019daggnndagstructurelearning}. Yet, recovering ground-truth structures remains difficult, especially in healthcare or data-scarce domains. Expert-driven approaches are viable but resource-intensive. Recent work \cite{kiciman2023causal} shows that LLMs, with their encoded world knowledge, can support causal reasoning and complement expert input.  

In this paper, we leverage multiple LLMs with different roles to mimic adversarial training in GAN and use the heuristic causal structure discovery to guide the data generation process. 

\section{Methods}
\subsection{Problem formulation}
Given a small labeled tabular dataset with $n$ instances and $d$ features, denoted as $D_{\text{real}} = {(\mathbf{x}, y)}$ where $\mathbf{x}$ represents a $d$-dimensional vector of features and $y$ indicates label. The features are described by natural-language strings like ``age'' or ``gender''. For synthetic data generation, we train a generator on a training subset $D_{\text{train}}$ of $D_{\text{real}}$, generating a synthetic dataset $D_{\text{syn}}$.
\subsection{Multi-agent LLM as GAN}
\begin{figure}
    \centering
    \includegraphics[width=\linewidth]{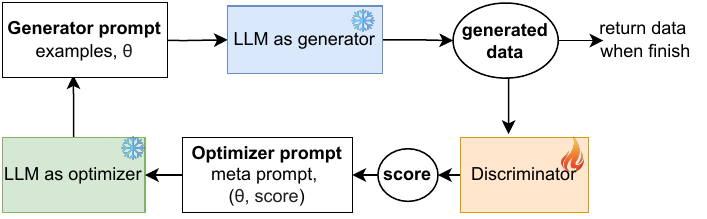} 
    \caption{Overview of MALLM-GAN. In each optimization step, the \colorbox{optimizer_color}{LLM as Optimizer} generates a data generation process $\theta$ in (\textbf{Generator prompt}) based on the pairs of previous $\theta$ and its score in \textbf{Optimizer prompt}. Then the \colorbox{generator_color}{LLM as Generator} uses the current $\theta$ and a few examples to generate data. We evaluate the $\theta$ using the accuracy (\textbf{score}) of \colorbox{discriminator_color}{Discriminator}. The more the data generation process $\theta$ is optimized, the lower the discriminator's accuracy. This adversarial optimization finishes when data generation process is no more improved.}
    \label{fig:overview}
\end{figure}

\textbf{Overview.} We propose to develop a multi-agent LLM as GAN (MALLM-GAN) that generates tabular data by mimicking adversarial optimization (Fig. \ref{fig:overview}).  The objective is to optimize the data generation process $\theta$, which is a natural language description of i) the problem description and ii) the simple data generation process or causal structures representing relationships between variables.  
In each iteration $i$,  an LLM agent \textbf{Generator} generates data $D_{\text{syn}}$ with $\theta_i$ and a batch in $D_{\text{train}}$; a supervised model \textbf{Discriminator} is accordingly optimized using $[D_{\text{train}}, D_{\text{syn}}]$ and evaluates $\theta_i$ using $D_{\text{test}}$; and another LLM agent \textbf{Optimizer} improves $\theta_i$ to decrease the discriminator's accuracy (Algorithm \ref{alg:pseudo_code}).  We repeat the iterations until the discriminator's accuracy converges or the iteration reaches the maximum epoch. 

\subsubsection{Generator}\label{method:generator}
\textbf{Data generation process.} 
The data generation process $\theta$ is described in natural language and prompts the generator LLM to create synthetic data. It includes: i) context of data collection, ii) data schema, iii) causal structure describing relationships between variables, and iv) task instruction. The context provides external knowledge on data collection (e.g., \textit{``this dataset includes subject's socioeconomic factors...''}). The data schema contains the meta-information of variables (e.g., name, description, type, and categorical values). These elements remain constant during optimization. The causal structure, represented as a DAG and converted into text format $(x_1, x_2)$, indicates $x_1$ causes $x_2$. Various serialization techniques were tested, but the original structured format proved most effective. The initial causal structure is heuristically determined (e.g., Hill climbing \cite{Tsamardinos2006}). The task instruction guides the goal, such as \textit{``produce accurate and convincing synthetic data''}. Through adversarial optimization, the causal structure and instructions are refined to reduce discriminator accuracy. Thus, for each iteration $i$, $\theta_i$ is:
\begin{align}
\theta_i &= [\text{context}][\text{schema}] \nonumber \\
         &\quad [\text{causal structure}]_i[\text{task instruction}]_i
\end{align}
Note that subscription for iteration $i$ will be omitted for simplicity without loss of generalizability. Also, note that we used the causal structure to convey the relationship between variables within the prompt; thus, obtaining the causal structure of the ground truth is not our primary goal. (An example of generator prompt is provided in Appendix Listing 1)

\textbf{Few shot examples.} 
The data generation process $\theta$ is supplemented with $n$ examples to leverage in-context few-shot learning. Structured data $(\mathbf{x}, y)$ is serialized into JSON format, e.g., \textit{\{age: 53, work class: self-emp, ...\}} (Detailed prompt can be found in the Appendix Section 1). Various natural language serializations were tested but had minimal impact on performance. The number $n$ of examples is crucial; a large $n$ allows learning from diverse examples, but may make LLM overlook instructions due to lengthy inputs; while a small $n$ avoids overflow, but under-utilizes the data. Our solution, ``batches in a batch,'' splits a batch into smaller pieces that fit the input token size, generates a set of synthetic data, and collates them into $D_{\text{\text{syn}}}$ (see Algorithm \ref{alg:pseudo_code} Line 6). This approach balances the trade-offs in in-context few-shot learning.

\textbf{LLM as generator.} 
The goal of the generator is to create text similar to but not identical to provided real samples, with the temperature parameter controlling the variability.The generator LLM runs multiple times with smaller examples in a batch, and the generated data is collated into $D_{\text{\text{syn}}}$. $D_{syn, i}$ denotes the synthetic data generated at iteration $i$. (An example is provided in Appendix Listing \ref{supp:generator_prmot}).

\subsubsection{Discriminator}
Based on the generated data, we evaluate and score the quality of $\theta$ by assessing how easy it is to distinguish generated synthetic data from real data. Naturally, this is supervised learning rather than a reasoning task with LLMs. We build a discriminator $f$ such that $f: \mathcal{X} \to c $ where $\mathbf{x} \in \mathcal{X}$ and $f(\mathbf{x})$ is the predicted label $c$, which is 1 if $\mathbf{x} \in D_{\text{train}}$ and 0 if $\mathbf{x} \in D_{\text{syn}}$.
Specifically, at each iteration $i$, a new set of synthetic data $D_{\text{syn}, i}$ is generated. We form the combined dataset $D_{\text{train}} \cup D_{\text{syn},i}$. We assign labels to the combined dataset by $D = \left\{ (\mathbf{x}, c) \mid \mathbf{x} \in D_{\text{train}}, c = 1 \right\}
\cup \left\{ (\mathbf{x}, c) \mid x \in D_{\text{\text{syn}, i}}, c = 0 \right\}$. We update the discriminator $f_i$ incrementally based on $f_{i-1}$. We evaluate the accuracy of the discriminator with $D_{\text{test}}$ and pass a pair of ($\theta_i$, $L(f_i)$) to the optimizer where $L(f)$ denotes the discriminatory power of $f$ (e.g., accuracy, likelihood). We prefer to use accuracy because this is a direct measurement we aim to increase and because our optimizer does not require numerical derivatives. 

The discriminator obtains better discriminatory accuracy to distinguish real or synthetic data as the discriminator accumulates the discriminatory power of past iterations $0, ...,i-1$ and is updated with newly generated, more realistic synthetic data from the current iteration $i$. However, on the other hand, as the $D_{\text{\text{syn}}}$ becomes more realistic over the iterations, it gets easier to fool the discriminator, and the discriminator's accuracy decreases. Therefore, our discriminator obtains better discriminatory power during this adversarial optimization. 

\subsubsection{Optimizer}
The parameter to be optimized is a text, $\theta$, which doesn't have derivatives. So we use optimization by prompting, which leverages LLM as an optimizer \cite{yang2024large}. To make LLM act as an optimizer, we provide a meta-prompt, which consists of two parts, the instruction and the prompt-score pairs $(\theta, L(f))$, An example is provided in Appendix Listing \ref{supp:optimizer_prompt}). 

To leverage LLM's in-context few-shot learning in the optimizer \cite{yang2024large}, we provide a few examples of possible solutions along with their scores from the discriminator's scores. Note that the example here is different from data $(\mathbf{x},y)$. We keep the top $k$ solution pairs over the past iteration as the optimization trajectory to guide the optimization. We sort the score, so that the more desirable $\theta$ goes to the end of prompt. This will allow the LLM to recognize patterns among the data generation process with better score. See examples in Appendix Listing \ref{supp:optimizer_output}.

A potential pitfall is that the discriminator score $L(f)$ from past iterations 
$0, \ldots, i-1$ is not directly comparable to the score at the current 
iteration $i$. Earlier discriminators $f_0, \ldots, f_{i-1}$ typically have 
weaker discriminative ability, which makes their reported scores unreliable 
for comparing parameter settings $\theta$ across iterations. To resolve this, 
we re-evaluate all past parameter candidates $\theta_0, \ldots, \theta_{i-1}$ using the current discriminator $f_i$. In 
other words, instead of relying on their originally recorded scores 
$L(f_0), \ldots, L(f_{i-1})$, we compute adjusted scores
\[
\tilde{L}_i(\theta_j) \;=\; L(f_i; \theta_j), \quad j < i,
\]
by passing the same candidate-generated samples through the latest discriminator. 
This ensures that all scores are measured against the most 
up-to-date discriminator, so that they are directly comparable when selecting 
the best $\theta$.

In total, the LLM optimizer takes as input the meta prompt and a series of data generation process $\theta$ and adjusted scores $L(f_i)$. The optimizer outputs the revised data generation process, particularly focusing on causal structure and task instruction. We repeat iterative optimization and generation until we reach the maximum iteration. 

\begin{algorithm}[tb]
\caption{Pseudocode for MALLM-GAN's optimization and generation}
\label{alg:pseudo_code}
\small
\begin{minipage}{\linewidth}  
\begin{algorithmic}[1]
    \STATE \textbf{Input:} $D_{\text{\text{train}}}$ (training data), $\theta$ (initial prompt), max\_epoch
    \STATE \textbf{Output:} Optimized prompt $\theta$
    
    \FOR{$i = 1$ to max\_epoch}
        \FOR{each batch in $D_{\text{\text{train}}}$}
            \STATE \textbf{Step 1: Run Generator}
            \STATE $D_{\text{\text{syn}}} \gets [\text{generator}(\theta + \text{example})$ for example in $D_{\text{\text{train}}}]$

            \STATE \textbf{Step 2: Run Discriminator}
            \STATE $c_x =1, x\in D_{real}; c_x=0, x\in D_{\text{\text{syn}}} $
            \STATE $(D, c)_{\text{\text{train}}}, (D, c)_{\text{\text{test}}}\gets split ((D_{\text{real}},c=1) (D_{\text{\text{syn}}}, c=0))$ 
            \STATE $\text{Discriminator.update}(D_{\text{\text{train}}}, c_{\text{\text{train}}})$
            \STATE $L \gets \text{Accuracy}(\text{Discriminator($D_{\text{test}}$)}, c_{\text{test}})$
            
            \STATE \textbf{Step 3: Run Optimizer}
            \STATE $\text{Pairs}_{(\theta, s)}.\text{append}((\theta, L))$
            \STATE $\theta \gets \text{Optimizer}(\text{instruction}+\text{Pairs}_{(\theta, L)})$
        \ENDFOR
    \ENDFOR
\end{algorithmic}
\end{minipage}
\end{algorithm}

\section{Experiments}\label{experiments}
\textbf{LLM backbone.} We used HIPAA-compliant Azure OpenAI GPT-4o\citep{gpt-4o} as our generator and optimizer. The generator's temperature was set to 0.5 to generate data points of highest confidence without randomly guessing, while the optimizer was set to be more creative using a temperature of 1. The top 3 prompt-score pairs is  the  to We also tried some open sourced LLM model, including Qwen3\cite{yang2025qwen3technicalreport} and LLama3\cite{grattafiori2024llama3herdmodels}, with different model sizes, to validate our framework's utility with different LLM's backbones. Open sourced models are deployed on one single NVIDIA H-100 80G.

Strong discriminators do not always contribute to a better generator \cite{arjovsky2017principledmethodstraininggenerative}. We tested Logistic regression, XGBoost, and neural network; we used the logistic regression model because it showed the highest performance while ensuring tractability during incremental updates over the iterations (Supplementary Figure 2).

Our benchmarks include several datasets from various domains: three public datasets (Adult\cite{misc_adult_2}, Medical Insurance\cite{med_ins}, Asia\cite{scutari2009learning}), and two private medical datasets (ATACH2, ERICH)  \cite{doi:10.1056/NEJMoa1603460,woo2013ethnic}. To ensure fair comparison without memorization concerns of LLM (e.g., public datasets are in the training corpus of LLM), private datasets were included. Details are in Supplement Table \ref{tab:data_description}.

We compare MALLM-GAN with multiple state-of-the-art tabular generative models such as traditional over-sampling techniques, SMOTE \cite{chawla2002smote}, the variational auto-encoder, TVAE \cite{xu2019Modeling}, the generative adversarial network, CTGAN \cite{xu2019Modeling}, the LLM-based synthetic data generation model, Be-GReaT\cite{borisov2023language}, and diffusion models, TabDDPM \cite{kotelnikov2023tabddpm} and Tabsyn \cite{zhang2024mixedtype}, the transformer-based tabular foundation model, TabPFN,\cite{hollmann_accurate_2025} to do generation task. Similar to MALLM-GAN, a prior work \cite{seedat2024curated} uses in-context few-shot learning of pre-trained LLMs but incorporates post-hoc data selection, which is beyond our scope. A comparison without post-hoc selection is available in Table \ref{table: components}. Specific hyper-parameters and computing resources are available in Supplement Section \ref{hyperparameters}.

We evaluated the impact of training data size $N=|D_{\text{train}}|$ on synthetic data quality by sampling subsets of different sizes ($N=25, 50, 100, 200$). We particularly aimed to compare performances in low and moderate data size. For fair comparison between real and synthetic data, synthetic data was generated to match the size of real data  ($|D_{train}|=|D_{syn}|$).
We held out 200 samples as the test set, and replicated experiments for each subsample five times to calculate the standard error of the evaluation metrics. 

\section{Results}\label{Results}

\subsection{Performance Evaluation}
We evaluate the performance of synthetic data generation models from two perspectives: Privacy leakage by Distance to Closest Records (DCR) and Machine Learning Efficiency (MLE) \cite{fang2024large, xu2019Modeling}.

\textbf{MLE.} To evaluate the utility of our synthetic data, we train supervised models on the synthetic datasets and assess their predictive performance on real test data ($D_{\text{test}}$). For classification tasks (Adult, Magic, Asia), we train logistic regression, random forest, Support Vector Machine, and XGBoost classifiers, reporting the $F1$ score. For regression tasks (Insurance, ATACH, ERICH), we train linear regression, random forest, and XGBoost regressors, reporting the coefficient of determination ($R^2$). For each setting, we average the best scores across random seeds. As a benchmark, we also train the same models using real data ($D_{\text{train}}$), which serves as the gold standard maximum likelihood estimate (MLE) for comparison.

As a result, MALLM-GAN generated high-quality synthetic tabular data across multiple datasets and training data sizes, outperforming baselines (Table \ref{table:MLE}), especially with a high dimension setting ($p>>n$, e.g. ATACH2 and ERICH). This indicates MALLM-GAN‘s robustness to smaller sample sizes, unlike baselines that require more data. While TabPFN also achieves comparable performance and scales well with increasing sample sizes, it has notable limitations—its effectiveness declines when the number of categorical levels exceeds 10 (e.g. Adult and ERICH) or when all variables are categorical (e.g. Asia), both of which are common scenarios in real-world datasets. Furthermore, MALLM-GAN outperformed the baselines in both public and private datasets, suggesting that it does not rely on the pre-trained LLM’s memorization. We also benchmark our model on the medium dataset scenario (N=400, 800). The results, as shown in Appendix Table \ref{tab-mle: medium}, demonstrate that the data-driven model can scale their performance as the dataset size increases, while our model can still achieve a comparable performance.

\textbf{DCR distributions.} 
The DCR metric assesses the realism and diversity of synthetic data. It determines whether the synthetic data points are too similar to the real data points (potential privacy leakage) or too dissimilar (hurting the utility of the synthetic data). The DCR is defined as $d(\mathbf{x}_{\text{syn}}, D_{\text{real}}) = \min_{\mathbf{x}_{\text{real}} \in D_{\text{real}}} l_1(\textbf{x}_{\text{syn}}, \textbf{x}_{\text{real}}).$\cite{borisov2023language} 

\begin{figure}[h]
    \centering
    \includegraphics[width=\linewidth]{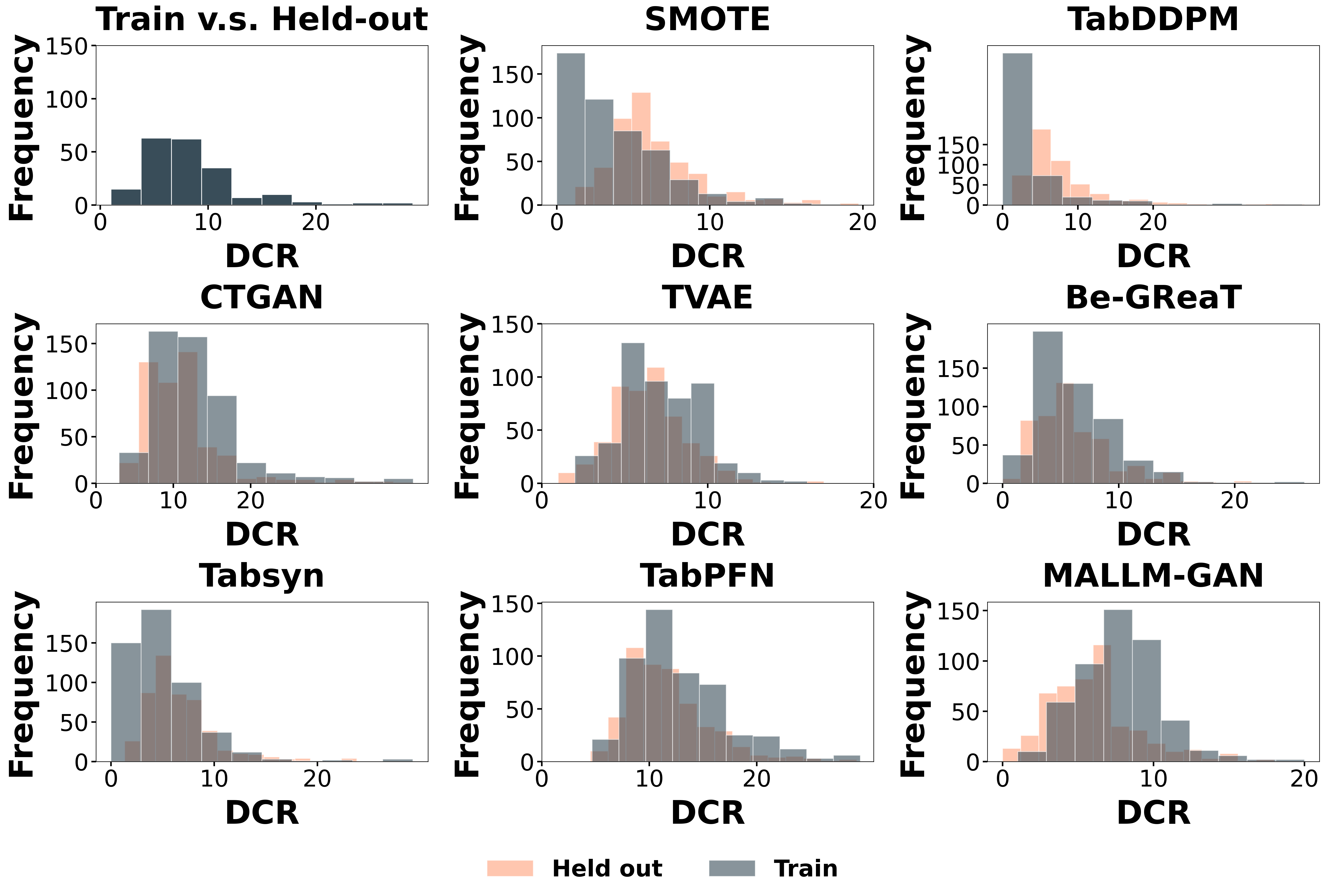}
    \caption{DCR between the synthetic data and the real data on Adult dataset. DCR was calculated based on training data and held-out test data for each model. A good model should have similar distributions between the DCR to training and the DCR to held-out dataset.}
    \label{fig:dcr plot}
\end{figure}

Figure \ref{fig:dcr plot} compares DCR distributions between train and held-out sets for various models on the Adult dataset (N = 100). While baseline models such as SMOTE and TabDDPM show overfitting on the training data, MALLM-GAN achieves the comparable DCRs compared with other baselines that generated samples closely match the real distribution while maintaining diversity, reflecting low memorization risk and strong generalization. These results highlight MALLM-GAN’s capability to generate realistic and privacy-preserving data even in small-sample settings, with consistent trends observed across other datasets (Fig. \ref{fig:dcr2}).

\begin{table*}[ht]
    \caption{Benchmark MLE results over 6 datasets. Baseline results were obtained from training the supervised models directly on the real data. SMOTE* interpolates data within the training set, thus it gets higher accuracy by copying training data and compromising DCR. "-" indicates that the model failed to generate.}
    \tiny
    \renewcommand{\arraystretch}{0.8}
     \begin{center}
     \begin{tabular}{@{}p{1cm}p{1.5cm}p{1.5cm}p{1.5cm}p{1.5cm}p{1.5cm}p{1.5cm}p{1.5cm}@{}}
     \toprule
     & & \multicolumn{4}{c}{Public dataset} & \multicolumn{2}{c}{Private dataset}\\
     & & Adult ($F1$) & Magic($F1$) & Asia ($F1$) & Insurance($R^2$) & ATACH($R^2$)& ERICH($R^2$)\\

     \cmidrule(lr){3-8}
     \multirow{5}{*}{N=25} & Real data & 0.80 &  0.79 & 0.83 & 0.52 & 0.25 & -0.23\\
       & SMOTE* & $0.82\pm0.02$ & $0.58\pm0.11$  & $0.83\pm 0.00$ &$0.80\pm0.01$ &  $0.06\pm0.20$ & $-0.15\pm0.12$ \\
       \cmidrule(lr){3-8}
       & CTGAN & $0.72\pm0.03$ & $0.60\pm 0.32$ & $0.75\pm0.10$ & $-0.31\pm0.29$& $-1.06\pm0.70$ & $-0.87\pm0.97$ \\
       & TVAE & $0.74\pm0.03$& $0.60\pm0.04$ & $\mathbf{0.84\pm0.03}$ & $-0.02\pm0.24$& $-0.01\pm0.07$&$-0.13\pm0.06$\\
       & Be-GReaT & $0.76\pm0.04$ & $0.71\pm 0.05$&$\mathbf{0.84\pm0.03}$ & $0.22\pm0.20$ & -&$-0.37\pm0.24$\\
       & TabDDPM & $0.61\pm0.12$& $0.58\pm0.11$& - & $-3.75\pm0.70$ & $-1.80\pm 1.24$ & $-1.65\pm0.05$\\
       & Tabsyn & $0.78\pm0.04$& $0.74\pm0.02$ & - & $-0.30\pm0.73$ & - &$-0.49\pm0.24$\\
       & TabPFN & $0.76\pm 0.01$& $0.69\pm0.04$  & - &$0.31\pm0.32$ & $0.01\pm0.29$& - \\
       \rowcolor{gray!30} 
       & \MyModel{}  & $\mathbf{0.82\pm0.03}$ & $\mathbf{0.79\pm0.00}$ & $0.74\pm0.01$ & $\mathbf{0.61\pm0.12}$ & $\mathbf{0.17\pm0.11}$& $\mathbf{-0.20\pm0.05}$\\
     \midrule
     \multirow{5}{*}{N=50} & Real data & 0.76 & 0.79 & 0.82 & 0.78 & 0.17 & -0.01 \\
       & SMOTE* & $0.74\pm0.00$ & $0.28\pm0.14$ & $0.83\pm 0.00$ &$0.80\pm0.01$ &  $0.20\pm0.11$ & $-0.10\pm0.10$ \\
       \cmidrule(lr){3-8}
       
       & CTGAN & $0.70\pm0.04$ & $0.64\pm 0.10$ & $0.41\pm0.18$ & $-0.48\pm0.58$& $-0.53\pm0.39$ & $-0.00\pm0.14$ \\
       & TVAE & $0.72\pm0.02$& $0.70\pm0.08$ &$\mathbf{0.82\pm0.01}$ & $0.37\pm0.23$& $-0.00\pm0.14$&$-0.11\pm0.12$\\
       & Be-GReaT & $0.71\pm0.03$ & $0.70\pm0.03$ & $0.80\pm0.04$ & $0.68\pm0.04$ & -&$-0.38\pm0.12$\\
       & TabDDPM & $0.59\pm0.15$ & $0.28\pm 0.14$ & - & $0.63\pm0.11$ & $-0.88\pm 0.71$ & $-1.02\pm0.21$\\
       & Tabsyn & $0.67\pm0.01$ & $0.73\pm0.03$ & - & $0.75\pm0.05$ & -&$-0.44\pm0.24$\\
       & TabPFN & $0.68\pm0.03$ & $0.66\pm 0.05$& - &$0.72\pm0.01$ & $\mathbf{0.22\pm0.07}$& - \\ 
       \rowcolor{gray!30}
       & \MyModel{}  & $\mathbf{0.74\pm0.01}$ & $\mathbf{0.76\pm0.01}$ & $\mathbf{0.82\pm0.00}$ & $\mathbf{0.77\pm0.03}$& $ 0.18\pm0.08$& $\mathbf{-0.07\pm0.07}$\\
     \midrule
      \multirow{5}{*}{N=100} & Real data & 0.86 & 0.83 & 0.83 & 0.82 & 0.26 & -0.04 \\
       & SMOTE* & $0.78\pm0.01$ & $0.78\pm0.05$& $0.83\pm 0.00$ &$0.80\pm0.01$ &  $0.27\pm0.03$ & $-0.15\pm0.13$ \\
       \cmidrule(lr){3-8}
       & CTGAN & $0.66\pm0.06$ & $0.62\pm 0.03$ & $0.63\pm0.19$ & $-0.09\pm0.11$& $-0.40\pm0.21$ & $-0.33\pm0.11$ \\
       & TVAE & $0.67\pm0.05$& $0.70\pm0.08$ &$\mathbf{0.83\pm0.01}$ & $0.39\pm0.15$& $-0.01\pm0.07$&$-0.11\pm0.12$\\
       & Be-GReaT & $0.76\pm0.04$ & $0.74\pm0.02$& $\mathbf{0.83\pm0.00}$ & $0.54\pm0.10$ & $-0.25\pm0.23$&$-0.38\pm0.12$\\
       & TabDDPM & $0.75\pm0.01$ & $0.78\pm 0.05$ & - & $-5.26\pm0.42$ & $-0.99\pm0.33$ & $-0.19\pm0.05$\\
       & Tabsyn & $0.75\pm0.02$ & $0.74\pm 0.02$ & - & $0.75\pm0.02$ & - & $-0.24\pm0.10$\\
       
     & TabPFN & - & $0.74\pm0.04$&- &$0.71\pm0.01$ & $\mathbf{0.32\pm0.05}$& - \\ 
     \rowcolor{gray!30} 
       & \MyModel{}  & $\mathbf{0.79\pm0.02}$ & $\mathbf{0.80\pm0.01}$  & $0.74\pm0.03$ & $\mathbf{0.72\pm0.00}$& $0.28\pm0.05$& $\mathbf{0.02\pm0.05}$\\
     \midrule
      \multirow{5}{*}{N=200} & Real data & 0.85 & 0.81 & 0.83 & 0.83 & 0.27 & 0.16\\
       & SMOTE*&$0.78\pm0.04$ & $0.77\pm0.02$ & $0.83\pm0.00$ & $0.79\pm0.02$ & $0.31\pm0.04$ & $0.05\pm0.06$\\ 
       \cmidrule(lr){3-8}
       & CTGAN & $0.61\pm0.02$& $0.59\pm0.05$ & $0.71\pm0.10$ & $-0.12\pm0.08$ & $-0.27\pm 0.05$ & $-0.19\pm0.10$\\
       & TVAE & $0.67\pm0.05$& $\mathbf{0.79\pm0.02}$&$0.82\pm0.01$& $0.62\pm0.05$ & $0.08\pm0.06$ & $-0.08\pm0.07$\\
       & BeGReaT & $0.69\pm0.05$ & $0.76\pm0.02$ & $0.82\pm0.00$ & $0.72\pm0.03$ & $0.16\pm0.06$ & $-0.18\pm0.16$\\
       & TabDDPM & $0.60\pm0.15$ & $0.77\pm0.02$ & - & $0.56\pm0.14$ & $-0.55\pm0.33$ & $-0.30\pm0.06$\\
       & Tabsyn & $0.72\pm0.08$ & $\mathbf{0.79\pm0.02}$ & - & $0.76\pm0.04$ & - & $ -0.14\pm0.07 $ \\
       & TabPFN & - & $0.69\pm0.03$ & - &$\mathbf{0.79\pm0.01}$ & $\mathbf{0.37\pm0.04}$ & - \\
       \rowcolor{gray!30} 
       & \MyModel{}  & $\mathbf{0.76\pm0.02}$ & $\mathbf{0.79\pm0.01}$ & $\mathbf{0.83\pm0.01}$ & $0.69\pm0.04$ & $\mathbf{0.37\pm0.06}$ & $\mathbf{0.02\pm0.02}$\\
     \midrule
     \end{tabular}
     \end{center}
     \label{table:MLE}
 \end{table*}

\subsection{Ablation study}
\textbf{Number $n$ of example in in-context few shot learning}. Due to the LLM's limited context length, we implemented a "batches in a batch" method to leverage all training data within these constraints (Section \ref{method:generator}). We varied the number $n$ of examples and found the optimal $n$ varied by different datasets considering their characteristics (Fig. \ref{fig:num-mle}). We varied the number $n$ of few-shot examples by 1-5 and measured the MLE (Fig. \ref{fig:num-mle}) and DCR distribution (Table \ref{table:num-shots-dcr1}, \ref{table:num-shots-dcr2}) to find the optimal number $n$. We found that simply increasing number of in-context samples does not necessary improve the quality of the generation. The MLE did not increase with more examples because more examples will increase the context length and the generator overlooks some key context information. And thus the optimal $n$ varies among different datasets given their heterogeneous complexity and domains.

\textbf{Causal structure and Optimization}. To assess the impact of each component on overall performance, we examined the contribution of the causal structure in the data generation process $\theta$ and the LLM as an optimizer. We compared the full model, which includes both components, to a version without them, similar to CLLM \cite{seedat2024curated} without post-processing data selection (Table \ref{table: components}). It shows that incorporating the causal structure alone does not improve the MLE compared to a model with only in-context few-shot learning. However, the LLM optimizer improved $\theta$ using prior knowledge encoded in LLM and finally achieved the highest MLE. Incorporating external knowledge into LLMs has been shown to significantly improve the quality of generated text, similar to retrieval-augmented generation (RAG) \cite{lewis2021retrievalaugmented}. Our approach shares this concept by incorporating a knowledge graph but optimizes the knowledge itself through adversarial optimization.

\textbf{Experiments on Different LLM backbones}
We further evaluate our framework using a variety of open-source LLM backbones, and the results are presented in Table \ref{tab:llmbackbone}. As shown, the framework performs consistently well across these relatively smaller models, achieving high MLE scores. This demonstrates the generalizability of our approach and aligns with the scaling law observation that larger and more capable LLMs tend to yield higher-quality synthetic data. Interestingly, we also observe that open-source models perform comparably to GPT-4o on public datasets but show a clear performance gap on more complex private datasets. This discrepancy may stem from potential data leakage or domain overlap in the training data of open-source models, or from the inherent limitations of smaller models when handling more complex data distributions.

\begin{table}[]
    \centering
    \scriptsize
    \renewcommand{\arraystretch}{0.8}
    \begin{tabular}{lccc}
    \toprule
    LLM Backbone & Model Size & Adult ($F1$) & ATACH ($R^2$)\\
    \midrule
     \rowcolor{gray!30}
     GPT-4o     & N/A   & $0.82\pm0.03$ & $0.17\pm 0.11$\\
     Qwen3-14B  & 14.8B & $0.82\pm0.02$ & $-0.13\pm 0.22$\\
     Qwen3-8B   & 8.19B  & $0.79\pm0.01$ & $-0.32\pm 0.10$\\
     Qwen3-4B   & 4.02B  & $0.78\pm0.05$& $-1.01\pm 0.10$\\
     Qwen3-1.7B & 2.03B  & $0.50\pm0.09$ & $-1.80\pm0.31$ \\
     LLama-3.1-8B-Instruct  & 8.03B   & $0.82 \pm 0.03$ & $0.07\pm 0.08$\\
     LLama-3.2-3B-Instruct & 3.21B & $0.58\pm0.10$& $-0.23 \pm 0.03$\\
     \bottomrule
    \end{tabular}
    \caption{Results with different LLM backbones on Adult dataset with N=25. Qwen3 were called without reasoning.}
    \label{tab:llmbackbone}
\end{table}

\begin{table}[h]
\centering
\caption{MLE of ablated models to evaluate the effects of causal structure in data generation process and optimization via LLM. Causal: Causal structure in data generation process, Opt: Optimization by LLM.}
\renewcommand{\arraystretch}{0.8}
\scriptsize
\begin{tabular}{@{}lccc@{}}
\toprule
   & \makecell{Few-shot} & \makecell{Few-shot\\+Causal} & \makecell{Few-shot\\+Causal\\+Opt (\MyModel{})}\\
   \cmidrule{2-4}
   Adult ($F1$) & $0.76\pm0.05$ & $0.75\pm0.04$ & $\mathbf{0.79\pm0.04}$ \\
   Asia ($F1$) & $0.23\pm0.00$ & $0.28\pm0.28$ & $\mathbf{0.83\pm0.00}$\\
   Insurance ($R^2$) & $0.68 \pm 0.02$  & $0.67\pm0.09$ & $\mathbf{0.72\pm0.05}$\\
   ATACH ($R^2$) & $0.16 \pm 0.09$  & $0.13\pm0.06$ & $\mathbf{0.27\pm0.07}$\\
   ERICH ($R^2$) & $-0.07\pm0.07$ & $\mathbf{0.03\pm0.04}$ & $-0.03\pm0.07$ \\
\bottomrule
\end{tabular}
\label{table: components}
\end{table}

\begin{figure*}[h]
    \centering
    \includegraphics[width=0.9\linewidth]{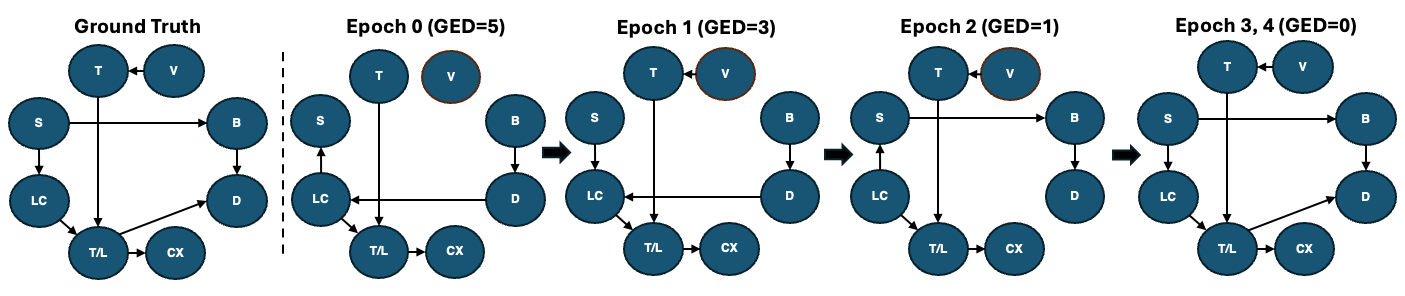}
    \caption{An example of trajectory of causal structure in data generation process over adversarial optimization using Asia dataset. \textbf{T}: Tuberculosis, \textbf{V}: Visit to Asia, \textbf{S}: Smoke, \textbf{LC}: Lung cancer, \textbf{T/L}: Tuberculosis or Lung cancer, \textbf{CX}: Chest X-ray, \textbf{D}: Dyspnea, \textbf{B}: Bronchitis. Graphical Edit Distance (GED) is used to measure the distance from the current generated causal graph to the ground truth.}
    \label{fig: asia_causal}
\end{figure*}

\begin{figure}[!h]
    \centering
    \includegraphics[width=\linewidth]{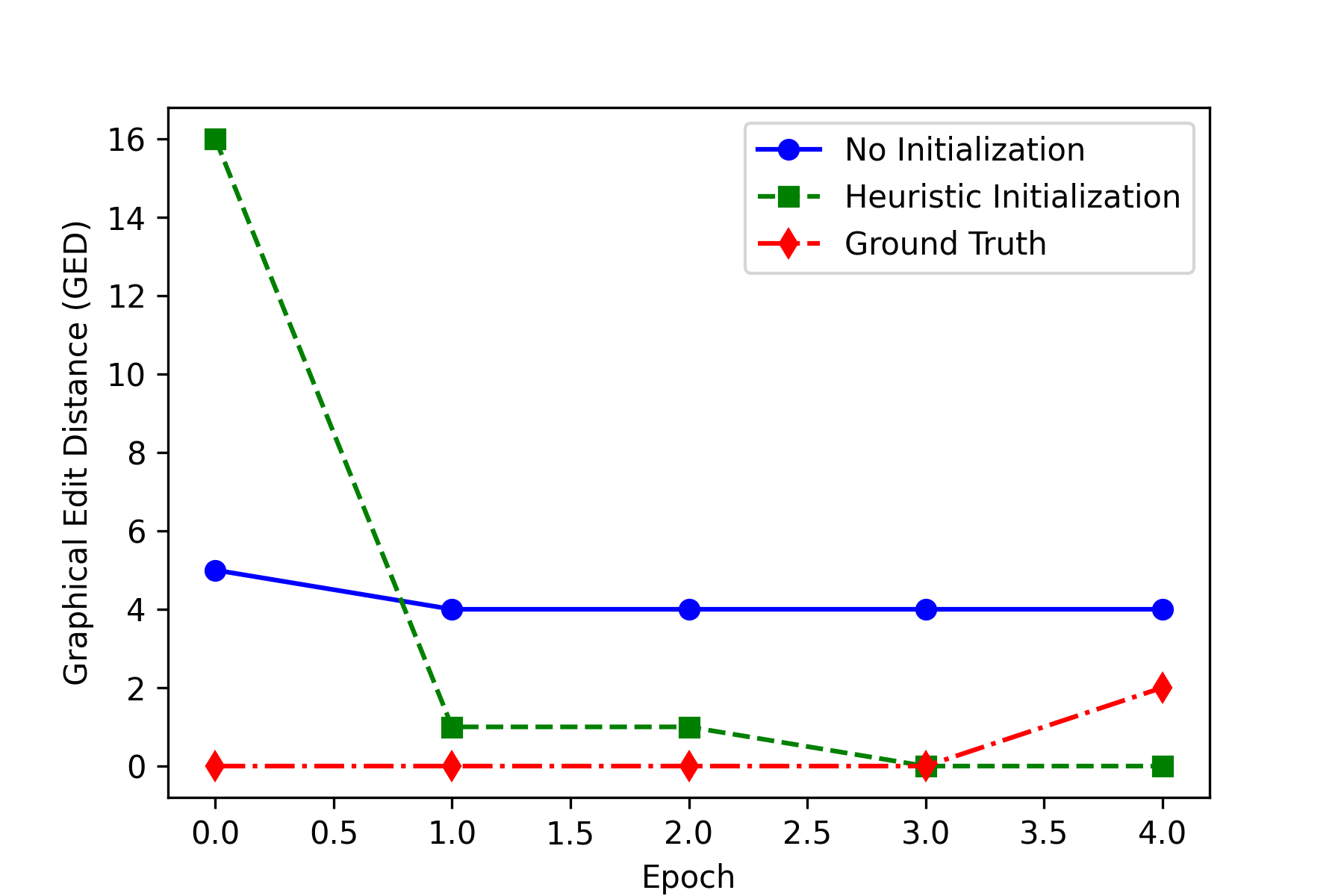}
    \caption{Causal structure initialized by different methods and the convergence step. Convergence is measured by GED to the ground truth causal graph. Experiment was conducted on the Asia dataset, which has the ground truth causal structure.}
    \label{CAG:GED}
\end{figure}

\textbf{Optimization trajectory of data generation process}
A key advantage of \MyModel{} is its transparent, text-described data generation process, which enables direct observation of how the generation mechanism evolves during adversarial optimization. Using the Asia dataset with known causal structures of ground truth, we visualized this trajectory: the learned causal graph progressively converges to ground truth (Fig.\ref{fig: asia_causal}), as reflected by decreasing GED values. Both heuristic and uninitialized structures showed convergence, driven by knowledge from the pre-trained LLM, though with distinct patterns (Fig.\ref{CAG:GED}). Moreover, Table~\ref{tab:asia_trajectory_task} discriminator accuracy declined over iterations with task instructions being increasingly specific, indicating that the synthetic data became more indistinguishable from real data.

\begin{table*}
    \centering
    \scriptsize
    \caption{Trajectory of task instruction in data generation process over adversarial optimization. Lower score is the better.}
    \begin{tabular}{c p{14cm} c}
        \hline
        \textbf{Iteration} & \textbf{Task instruction} & \textbf{Score \textdownarrow} \\
        \hline
        Epoch 1 & \textit{``The ultimate goal is to produce accurate and convincing synthetic
data that dutifully represents these causal relationships given the user provided samples.''} & 100.0\% \\
        \hline
        Epoch 2 & \textit{``The ultimate goal is to create a detailed and convincing dataset that accurately mirrors these causal pathways. While synthesizing your data, keep in mind the following key relationships: a 'visit to Asia' increases the likelihood of 'tuberculosis', 'smoking' can lead to 'lung cancer' and 'bronchitis', and both 'tuberculosis' and 'lung cancer' can contribute to 'either tuberculosis or lung cancer', which in turn can lead to 'Dyspnea'. Also, take note of how both 'tuberculosis' and 'lung cancer' are associated with 'chest X-ray' results. Your data should reflect these intricate relationships while remaining consistent and realistic.''} & 76.19\% \\
        \hline
        Epoch 4 & \textit{``You are tasked with generating a synthetic dataset that faithfully demonstrates the given causal connections. Make sure the dataset illustrates how a 'visit to Asia' can cause 'tuberculosis', how 'smoking' can lead to 'lung cancer' and 'bronchitis', and how either 'tuberculosis' or 'lung cancer' can eventually incite 'Dyspnea'. Also, the dataset should reasonably reveal how a 'chest X-ray' ties in with 'tuberculosis' and 'lung cancer'. Ensure the synthetic data reflects realistic scenarios where these factors interact, affecting each other exactly as per these defined causal relationships.''} & 66.67\% \\
        \hline
    \end{tabular}%
    \label{tab:asia_trajectory_task}
\end{table*}

\begin{figure}[!h]
    \centering
    \includegraphics[width=0.9\linewidth]{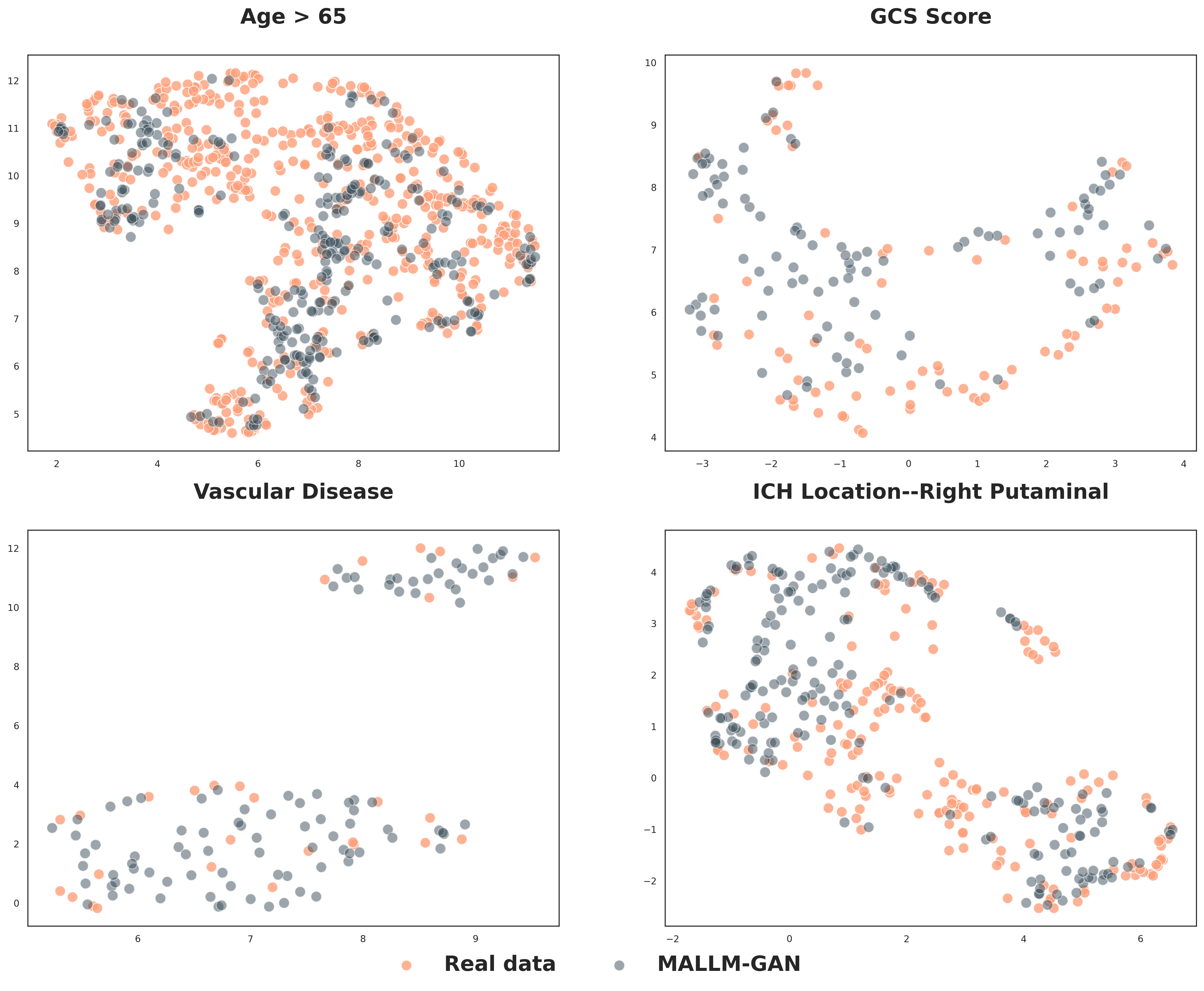}
    \caption{Real and synthetic data distribution with three categorical conditions and one numerical range condition in ERICH data.}
    \label{fig:rare_conditions}
\end{figure}

\textbf{Conditional sampling.} We leverage the generator’s conditional capability to synthesize data under user-defined constraints on categorical values and numerical ranges, comparing \MyModel{} with baseline models via UMAP visualization. For categorical conditions, we selected three rare subgroups in the ERICH dataset—(i) \textit{hematoma location} = \textit{right putaminal}, (ii) \textit{GCS score} = 13, and (iii) \textit{prior vascular disease}—with 187, 83, and 29 patients, respectively. Baseline models failed to generate realistic samples due to limited data, whereas \MyModel{} produced distributions closely matching the real data (Fig. \ref{fig:rare_conditions}). For the numeric condition \textit{Age} $> 65$ (534 patients), baselines could not model range-based constraints, but \MyModel{} successfully generated condition-consistent data, demonstrating flexible comprehension of natural-language conditions.

\section{Conclusions}
We propose a novel framework to generate synthetic tabular data by leveraging multiple LLMs to address the data scarcity issue. Compared with other LLM-based methods, the in-context learning approach does not require fine-tuning on LLM but still leverages the whole data. We demonstrate that LLM can help generate high-quality data for downstream tasks with an optimized prompt of domain knowledge and it enables transparent data generation with interpretation. 


\section{Limitations}
Our proposed framework has several limitations. First, due to the limited context length of current large language models (LLMs), the method does not scale well to extremely high-dimensional datasets. As the number of features increases, the contextual input becomes excessively long, which may degrade generation quality and reduce the reliability of synthetic data. Although future LLMs with extended context capabilities may alleviate this issue, it remains a practical constraint in the current setting. Second, LLMs are known to struggle with high-quality random number generation \cite{hopkins2023can}, which can negatively affect the fidelity of synthetic data, particularly for datasets with many continuous variables where accurate stochasticity is essential. Third, similar to generative adversarial frameworks, our method lacks theoretical guarantees on convergence, which may lead to instability during training and sensitivity to hyperparameter choices. Moreover, while the proposed approach demonstrates clear advantages in low-data regimes, its relative performance gains diminish as the dataset size increases, suggesting limited benefits in large-scale settings. In addition, both training and generation incur non-trivial computational costs, especially for large datasets. Finally, although synthetic data generation is often considered a privacy-preserving alternative, it does not inherently prevent privacy leakage. In particular, adapting membership inference and attribute inference attacks to small-sample, mixed-type tabular data generated by LLM-based models remains an important open problem.

\section{Ethical Considerations}
Beyond the methodological limitations, we recognize several ethical risks associated with our approach. First, the optimization process in our model does not guarantee convergence, and the resulting prompts may reflect biases inherited from the pre-trained LLM backbone. This means that any optimized prompt or generated content should be interpreted cautiously, as it may be influenced by spurious correlations or latent biases in the underlying language model. Also, since our method does not learn the true data distribution of the source domain, synthetic samples generated by the model may not be statistically representative. Consequently, these data are unsuitable for inferential analyses or for drawing causal conclusions about real-world phenomena. They should be used primarily for model benchmarking or methodological exploration rather than for policy or decision-making.

\bibliography{custom}

\appendix

\section{Appendix}
\section{Prompt Examples}
Here, we provided examples of generator prompts and optimizer prompts. Note that the generator prompt evolves over the iterations. 

\begin{listing*}

\begin{lstlisting}[caption=Example of generator prompt,label=supp:generator_prmot, escapeinside=``]
System role: 
% Specify role and task
You are a data generation model. Your task is to understand the instruction below and generate tabular data.

% Context of data
<context>The dataset include subject`s social economic factors and demographics with the label that indicates whether their income is higher than 50k. </context>

% Data schema
<schema> age (numerical),workclass (categorical), education (categorical), education-num (numerical),marital-status (categorical), occupation (categorical), relationship (categorical), race (categorical), sex (categorical), capital-gain (numerical),capital-loss (numerical),hours-per-week (numerical),native-country (categorical), Income (categorical) </schema>

%Categorical variables and their available categories
<categorical variables>  workclass: {'Private', 'Local-gov', 'Without-pay', 'Self-emp-not-inc', 'State-gov', 'Federal-gov', 'Self-emp-inc'}, education: {'Some-college', 'Masters', '11th', '1st-4th', '7th-8th', 'Bachelors', 'Doctorate', '12th', '5th-6th', 'Prof-school', 'Assoc-voc', 'Assoc-acdm', '10th', '9th', 'HS-grad'}, marital-status: {'Divorced', 'Married-spouse-absent', 'Married-civ-spouse', 'Never-married', 'Widowed', 'Separated'}, occupation: {'Handlers-cleaners', 'Transport-moving', 'Sales', 'Prof-specialty', 'Farming-fishing', 'Machine-op-inspct', 'Adm-clerical', 'Other-service', 'Craft-repair', 'Protective-serv', 'Exec-managerial', 'Tech-support', 'Priv-house-serv'}, relationship: {'Wife', 'Not-in-family', 'Other-relative', 'Unmarried', 'Own-child', 'Husband'}, race: {'Black', 'Amer-Indian-Eskimo', 'Other', 'Asian-Pac-Islander', 'White'}, sex: {'Male', 'Female'}, native-country: {'Vietnam', 'Mexico', 'Hong', 'Taiwan', 'Italy', 'Portugal', 'Ireland', 'Guatemala', 'El-Salvador', 'United-States'}, Income: {'>50K', '<=50K'}
</categorical variables>

%causal structure
<causal structure> Consider this optimized causal graph of the data, where a pair (A, B) is used to represent a scenario where A affects B: [('age', 'workclass'), ('education', 'education-num'), ('education-num', 'Income'), ('marital-status', 'relationship'), ('occupation', 'Income'), ('hours-per-week', 'Income'), ('workclass', 'Income')]

This adjusted graph introduces 'education-num', which is a key determinant of 'Income'. Be sure to reflect 'age' impact on 'workclass' and 'marital-status' effect on 'relationship'. When creating the 'Income' data, pay careful attention to the roles of 'education', 'education-num', 'occupation', and 'hours-per-week' as stated in the causal graph.
</causal structure>

%Task
<task> The ultimate goal is to produce accurate and convincing synthetic data that dutifully represents these causal relationships. As such, strive for a quality score that is less than 70.0%. </task>



User role:     
% Example
<example> Here are examples from real data: 
[{'age': 53.0, 'workclass': 'Self-emp-not-inc', 'education': '10th', 'education-num': 6.0, 'marital-status': 'Married-civ-spouse', 'occupation': 'Farming-fishing', 'relationship': 'Husband', 'race': 'White', 'sex': 'Male', 'capital-gain': 0.0, 'capital-loss': 0.0, 'hours-per-week': 60.0, 'native-country': 'United-States', 'Income': '<=50K'}, {'age': 23.0, 'workclass': 'Private', 'education': 'HS-grad', 'education-num': 9.0, 'marital-status': 'Never-married', 'occupation': 'Adm-clerical', 'relationship': 'Own-child', 'race': 'White', 'sex': 'Female', 'capital-gain': 0.0, 'capital-loss': 0.0, 'hours-per-week': 40.0, 'native-country': 'United-States', 'Income': '<=50K'}]
</example>

<instruction>
Generate two synthetic samples mimic the provided samples. DO NOT COPY the samples and try to make the generated samples diverse. The response should be formatted strictly as a list in JSON format, suitable for direct use in data processing scripts such as conversion to a DataFrame in Python. No additional text or numbers should precede the JSON data.
</instruction>  
\end{lstlisting}

\end{listing*}

\begin{listing*}
    \begin{lstlisting}[language=Python, basicstyle=\ttfamily\footnotesize, caption=Example of input real example, label=supp:generator_output]
json
[{"treatment": 0, "age": 68.2, "ICH volume": 4.1, "ICH Location": "L Lobar", "IVH volume": 0.2, "GCS score": 14.3, "NIHSS score": 11.7, "Systolic blood pressure": 195.0, "Diastolic Blood Pressure": 83.0, "Hypertension": 1, "Hyperlipidemia": 1, "Type I Diabetes": 0, "Type II Diabetes": 0, "Congestive heart failure": 0, "Atrial Fibrillation": 0, "PTCA": 0, "Peripheral Vascular Disease": 0, "Myocardial fraction": 0, "Anti-diabetic": 0, "Antihypertensives": 1, "White blood count": 4.3, "Hemoglobin": 12.5, "Hematocrit": 37.7, "Platelet count": 129.0, "APTT": 35.3, "INR": 1.1, "Glucose": 148.0, "Sodium": 145.0, "Potassium": 4.1, "Chloride": 106.0, "CD": 30.1, "Blood urea nitrogen": 18.0, "Creatinine": 1.2, "race": "White", "sex": "Female", "ethnicity": "Hispanic", "mRS score after 30 days": 2.7}]
\end{lstlisting}
\end{listing*}

\begin{listing*}
\begin{lstlisting}[caption=Example of optimizer prompt, label=supp:optimizer_prompt]
System role: 
% Specify role and task
Your task is to optimize prompts for generating high-quality synthetic data. Aim to lower the scores associated with each casual structure and prompt, where a lower score reflects better quality. Here are the steps:
1. Examine the existing prompt-score pairs.
2. Adjust the causal graph to better represent the underlying relationships by adding or removing connections, and consider incorporating new features from the list {self.cols}.
3. Modify the prompt guidance to align with the revised causal graph, ensuring it aids in reducing the score. 

User role:
<pair>
Reflecting the adjusted causal graph of the data, where each tuple (A, B) indicates that A impacts B:
[('age', 'workclass'), ('marital-status', 'relationship'), ('marital-status', 'Income'), ('relationship', 'sex'), ('education', 'Income'), ('occupation', 'Income'), ('workclass', 'Income'), ('hours-per-week', 'Income')]
 
Use this causal graph as a guide to generate synthetic data that closely mirrors the real-world dataset. Remember to factor in the influence of 'age' on 'workclass', and 'marital-status' on 'relationship' and 'Income'. The 'relationship' should guide the generation of the 'sex' attribute. Further, take into consideration the effects of 'education', 'occupation', and 'hours-per-week' on 'Income' when synthesizing your data. The goal is to produce synthetic data that convincingly mimic these causal relationships.
Set your aim to achieve a score below 75.0%.
Score: 80.0%
</pair>
 
<pair>
Consider the revised and detailed causal graph of the data, which includes ('age', 'workclass'), ('marital-status', 'relationship'), ('relationship', 'sex'), ('education', 'Income'), ('occupation', 'Income'), ('workclass', 'Income'), ('hours-per-week', 'Income'):
 
In light of the causal graph, generate synthetic samples that mimic the structure in the provided dataset. Values such as 'age' should reflect on 'workclass'; 'marital-status' and 'relationship' should collaborate to inform 'sex', while 'education', 'occupation', 'workclass', and 'hours-per-week' should exhibit their influence on 'Income'. Also consider 'marital-status' influence on 'Income'. Your aim is to generate synthetic data that fully embody the interconnections within this causal graph.
Aim to achieve a score lower than 75%
Score: 80.95%
</pair>

<pair>
Here is the causal graph of the data, where a tuple (A, B) indicates A causes B:
[('marital-status', 'relationship'), ('marital-status', 'Income'), ('relationship', 'sex')]
Given the description of the data, generate synthetic samples that mimic the provided samples. 
Score: 85.71%
</pair>

Your updated prompt should explicitly include any modifications to the causal graph and guidance. The aim is to create a prompt that leads to the lowest possible score.
 
The updated prompt:
\end{lstlisting}

\end{listing*}

\begin{listing*}
 \begin{lstlisting}[language=Python, caption= Example of optimizer output, label=supp:optimizer_output]
<Causal structure> The optimized causal network, suggesting the influence of variable A on variable B, includes the following relationships: [('Age', 'Hyperlipidemia'), ('Hyperlipidemia', 'Type II Diabetes'), ('Type II Diabetes', 'Blood urea nitrogen'), ('Blood urea nitrogen', 'Creatinine'), ('Hypertension', 'Congestive heart failure'), ('Congestive heart failure', 'Atrial Fibrillation'), ('Atrial Fibrillation', 'GCS score'), ('GCS score', 'mRS score after 30 days'), ('Anti-diabetic', 'Type I Diabetes'), ('Type I Diabetes', 'Antihypertensives'), ('Antihypertensives', 'Potassium'), ('Potassium', 'Sodium'), ('PTCA', 'Peripheral Vascular Disease'), ('Peripheral Vascular Disease', 'Myocardial fraction'), ('Myocardial fraction', 'Hemoglobin'), ('Hemoglobin', 'Hematocrit'), ('race', 'ethnicity'), ('Sex', 'Hyperlipidemia')]</Causal structure>
 
<Task> Your task is to create realistic synthetic patient data, keeping the altered causal relationships as your guiding principle. Ensure the data reflects a diverse set of potential patient scenarios, evidencing the variety of health conditions one might find in a clinical setting. Remember that the engineered data should present unique, individual patient scenarios, each portraying a different, complex clinical situation. The synthetic data needs to be representative of different demographics ('Sex', 'race', 'ethnicity') and should also take into consideration different health conditions and treatment plans.</Task>"
    
\end{lstlisting}
\end{listing*}

\begin{listing*}
    \begin{lstlisting}[language=Python, caption= Modified instruction in generator prompt for conditional sampling, label=supp:generator_conditional]
<Instruction> 
Generate {number of samples in real data meeting the conditions} synthetic samples with {user-provided conditions}. Response should be formatted strictly as a list in JSON format, suitable for direct use in data processing scripts such as conversion to a DataFrame in Python. No additional text or numbers should precede the JSON data.
</Instruction>
\end{lstlisting}
\end{listing*}

\section{Experiment details}\label{Supp:experiment}
\subsection{Benchmark datasets descriptions}
We provide a detailed description of the benchmark data in Table \ref{tab:data_description}. All the public data are licensed under CC BY-4.0. The two private datasets (ATACH2 and ERICH) are available by proper request to NIH. The private datasets have been de-identified before releasing to us for research purpose. 

All the texts in the dataset, including data summary, headers, and categorical variables recorded in strings, are in English. 
\begin{table*}
    \centering
     \begin{tabular}{@{}l p{2.5cm} p{1.5cm} p{7cm} p{2cm} @{}}
     \toprule
         &  \# samples&  \# features & Description & Source\\
         \cmidrule(lr){2-5}
         Adult & 32,561 & 14 & The dataset includes people's socioeconomic factors and demographics, with the label that indicates whether their income is higher than 50k. & \cite{misc_adult_2}\\
         \cmidrule(lr){2-5}
         Magic & 19,020 & 10 & It is a simulated registration of high-energy gamma particles in a ground-based atmospheric Cherenkov gamma telescope using the imaging technique & \cite{magic_gamma_telescope_159} \\
         \cmidrule(lr){2-5}
         Medical Insurance & 2,772 & 7 & This dataset describes the paitents' demographics with their health insurance bills. & \cite{med_ins} \\
         \cmidrule(lr){2-5}
         Asia & 10000 & 8 & This is the dataset used to illustrate the utility of Baysian network to do causal structure discovery. The dataset is available in the R-package.  & \cite{scutari2009learning}\\
         \cmidrule(lr){2-5}
         ATACH2 & 1,000 & 37 & This is an RCT data that investigate in treatment for Intracerebral hemorrhage patients.&\cite{doi:10.1056/NEJMoa1603460}\\
         \cmidrule(lr){2-5}
         ERICH & 1,521 & 29 & The data is from a case-control study of Intracerebral Hemorrhage study which aims to investigate in the Ethnic/Racial variations. & \cite{woo2013ethnic}\\
         \bottomrule
    \end{tabular}
    \caption{Datasets description}
    \label{tab:data_description}
\end{table*}

\subsection{Hyperparameters} \label{hyperparameters}
Specific hyperparameters for each model are provided below.
\begin{itemize}
    \item \textbf{CTGAN}: Default parameters
    \item \textbf{TVAE}: Default parameters
    \item \textbf{BeGReaT}:
        \begin{itemize}
            \item Base LLM: Distiled-GPT2
            \item Batch size: 40
            \item Epochs: Depend on the feature numbers and the total sample size. (200-400)
        \end{itemize}
    \item \textbf{MALLM-GAN}:
    \begin{itemize}
        \item Temperature for generator: 0.5
        \item Temperature for optimizer: 1.0
        \item Batch size: 50
        \item Discriminator: XGBoost (max depth: 3, eta: 0.3, objective: binary:logistic)
    \end{itemize}
    \item \textbf{TabDDPM}: Default parameters
    \item \textbf{Tabsyn}: Default parameters
\end{itemize}

\section{Additional Experiments on Medium Size Datasets}
To evaluate our model's performance scaling with larger sample sizes, We also benchmark our model on the datasets of medium sample sizes(N=400, 800). The results are shown in Table \ref{tab-mle: medium}.
\begin{table*}
   \scriptsize
    \renewcommand{\arraystretch}{0.8}
     \centering
     \begin{tabular}{@{}p{0.6cm}p{2.0cm}p{1.5cm}p{1.5cm}p{1.5cm}p{1.5cm}p{1.5cm}p{1.5cm}}
    
     \toprule
     & & \multicolumn{3}{c}{Public dataset} & \multicolumn{2}{c}{Private dataset}\\
     & & Adult ($F1$) & Magic($F1$) & Asia ($F1$) & Insurance($R^2$) & ATACH($R^2$)& ERICH($R^2$)\\

     \cmidrule(lr){3-8}
     \midrule
      \multirow{5}{*}{N=400} & Real data & 0.83 & 0.82 & 0.84 &  0.85 & 0.31 & 0.18\\
       & SMOTE* & $0.85\pm0.03$  & $0.81\pm0.01$  & $0.84\pm0.00$ & $0.83\pm0.00$& $0.32\pm0.02$& $0.07\pm0.05$ \\
       \cmidrule(lr){3-7}
       & TabDDPM & $\mathbf{0.82\pm0.03}$ & $\mathbf{0.81\pm 0.01}$ & - & $\mathbf{0.79\pm0.03}$ & $\mathbf{0.36\pm0.02}$ & $\mathbf{0.09\pm0.04}$\\
       & CTGAN & $ 0.63\pm0.02$ & $0.60\pm0.02$ &  $0.59\pm0.17$ & $-0.18\pm0.10$ & $-0.08\pm0.07$ & $-0.24\pm0.10$\\
       & TVAE & $0.71\pm0.07$ & $0.79\pm0.01$  & $0.71\pm0.07$ & $0.62\pm0.05$ & $0.16\pm0.08$ & $-0.19\pm0.06$\\
       & Be-GReaT & $0.79\pm0.04$ & $0.79\pm0.02$&  $0.79\pm0.00$ & $0.72\pm0.03$ & $0.20\pm0.06$&$-0.13\pm0.07$\\
       & Tabsyn &$\mathbf{0.83\pm0.02}$ & $0.80\pm0.01$ & - &$\mathbf{0.82\pm0.02}$ & $\mathbf{0.40\pm0.04}$& - \\
       & TabPFN & - &  $0.77\pm0.02$ & - &$0.71\pm0.01$ & $\mathbf{0.40\pm0.04}$& - \\
       & MALLM-GAN & $0.79\pm0.02$ & $0.80\pm0.01$ & $\mathbf{0.83\pm0.00}$ & $0.71\pm0.03$ & $0.27\pm0.04$ & $0.02\pm0.03$\\
     \midrule
      \multirow{5}{*}{N=800} & Real data & 0.71 & 0.81 & 0.84 & 0.85 & 0.40 & 0.21\\
       & SMOTE* & $0.71\pm0.03$& $0.82\pm0.01$&$0.84\pm0.00$& $0.83\pm0.00$ & $0.37\pm0.03$& $0.10\pm0.05$\\ 
       \cmidrule(lr){3-8}
       & TabDDPM & $0.70\pm0.03$ & $0.82\pm 0.01$&  - & $\mathbf{0.83\pm0.01}$ & $-0.53\pm0.45$ & $\mathbf{0.12\pm0.04}$\\
       & CTGAN & $0.64\pm0.05$ & $0.54\pm0.05$ & $0.48\pm0.06$ & $-0.41\pm0.06$ & $-0.05\pm0.06$ & $-0.04\pm0.02$\\
       & TVAE & $0.77\pm0.02$ & $0.78\pm0.01$&  $0.82\pm0.01$ & $0.68\pm0.01$ & $0.12\pm0.07$ & $-0.05\pm0.03$\\
       & Be-GReaT & $0.75\pm0.07$ & $0.75\pm0.01$& $0.82\pm0.00$ & $0.53\pm0.21$ & $0.00\pm0.07$ & $-0.04\pm0.05$\\
       & Tabsyn & $0.78\pm0.08$ & $0.82\pm0.01$& - &$\mathbf{0.82\pm0.01}$ & $\mathbf{0.42\pm0.04}$ & - \\
       & TabPFN & - & - & - &$0.71\pm0.01$ & $0.40\pm0.04$ & - \\
       
       & MALLM-GAN& $\mathbf{0.80\pm0.02}$ & $0.79\pm0.00$& $\mathbf{0.84\pm0.00}$ & $0.72\pm0.01$ & $\mathbf{0.36\pm0.02}$ & $0.02\pm0.02$\\
       \midrule
    \end{tabular}
     \caption{Experiments on sample size 400, 800}
    \label{tab-mle: medium}
\end{table*}

\subsection{DCR evaluation on other datasets}
The following figures are to evaluate DCR on other datasets:
\begin{figure*}
\centering
    \subfigure[Insurance]{
        \includegraphics[width=1.0\textwidth]{sections/images/ins_dcr.png}
    }
    \subfigure[Magic]{
        \includegraphics[width=1.0\textwidth]{sections/images/magic_dcr.png}
    }
    \subfigure[ATACH2]{
        \includegraphics[width=1.0\textwidth]{sections/images/atach_dcr.png}
    }
    \subfigure[ERICH]{
        \includegraphics[width=1.0\textwidth]{sections/images/erich_dcr.png}
    }
\caption{DCR evaluation results.}

\label{fig:dcr2}
\end{figure*}

\section{Comparison of different discriminators}
In the study, we compare 3 different kinds of supervised classification models as the role of a discriminator. An experiment was conducted on the Adult dataset's sub-sample to demonstrate the discriminator's effects on the quality of the generated data. 
\begin{table*}
        \begin{center}
            \begin{tabular}{@{}lcccc@{}}
            \toprule
              & N = 100 & N = 200 & N = 400 & N = 800\\
            \cmidrule(lr){2-5}
            XGBoost & $0.78\pm0.03$ & $0.73\pm0.01$ & $0.76\pm 0.06$ & $0.72\pm0.00$\\
            Logistic regression  & $0.79\pm 0.02$ & $\mathbf{0.77 \pm 0.02}$ & $
            \mathbf{0.79\pm0.03}$ & $\mathbf{0.80\pm0.02}$ \\
            Neural Network & $\mathbf{0.80\pm0.02}$ & $0.57\pm0.12$ & $0.78\pm0.06$ & $0.67\pm0.12$\\
        \bottomrule
        \end{tabular}
        \end{center}
        \caption{Comparison of different discriminators' effects on the quality of the synthetic data. An experiment on sub-sample of Adult data.}
        \label{tab: discriminator}
\end{table*}

\section{Computing resource details}
The model proposed in this study does not require extensive computing resources for fine-tuning. However, this model requires access to the Azure service. For other baseline models, they are implemented on an NVIDIA A100 40GB GPU.

\section{Ablations studies on the number of provided real samples}
We conducted experiments to test the number of real examples' effects on the downstream evaluation metrics. Table \ref{table:num-shots-dcr1} and Table \ref{table:num-shots-dcr2} shows the DCR distance to the training and testing datasets respectively. We can learn from the tables that there is no association between the number of examples and the DCR. 

\begin{figure*}
    \centering
    \includegraphics[width=0.8\textwidth]{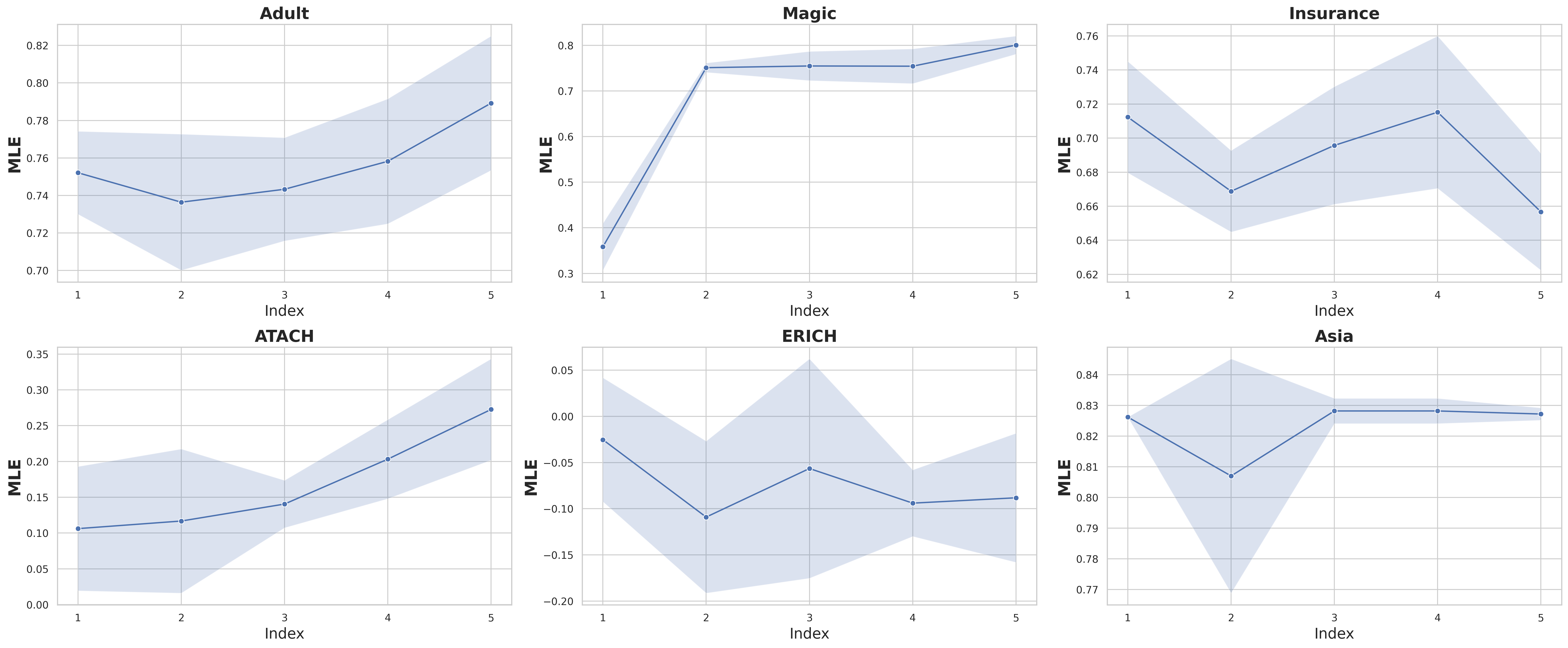}
    \caption{Number $n$ of examples and MLE.}
    \label{fig:num-mle}
\end{figure*}

Also, the Figure \ref{fig:num-mle} shows the MLE efficacy given different in-context number of shots. We can see that the patterns differentiate among different datasets. It is because that the complexity of the data can affect the context length and thus cast effect on the generation quality of the data.

\begin{table*}[ht]
        \centering
        \begin{tabular}{lccccc}
            \toprule
            & 1 & 2 & 3 & 4 & 5\\
            \cmidrule(lr){2-6}
            Adult & 5, 6, 10 & 5, 7, 12 & 4, 6, 9 & 4, 6, 10 & 4, 6, 11\\
            Magic & 23, 29, 37 & 24, 33, 52& 30, 44, 65 & 37, 53, 72 & 40, 55, 73\\ 
            Insurance & 31, 93, 301 & 44, 66, 453 & 32,60, 182 & 33, 73, 167 & 29, 55, 168\\
            ATACH2 & 61, 73, 88 & 72, 87, 97 & 69, 78, 89 & 66, 75, 94 & 67, 83, 104\\
            ERICH & 53, 61, 79& 59, 78, 96 & 59, 74, 101 & 56, 72, 96 & 50, 64, 88\\
            \bottomrule
        \end{tabular}
        \caption{Number $n$ of examples and DCR to \textbf{
        training dataset}. 25\%, 50\% (Median), 75\% quantile.}
        \label{table:num-shots-dcr1}
\end{table*}

\begin{table*}[ht]
        
        \begin{center}
            \begin{tabular}{lccccc}
            \toprule
            & 1 & 2 & 3 & 4 & 5\\
            \cmidrule(lr){2-6}
            Adult & 4, 7, 10& 5, 7, 11 & 5, 6, 10 & 4, 7, 10 & 4, 7, 11\\
            Insurance & 30, 115, 337 & 34, 91, 405 & 36, 76, 245 & 24, 64, 170 & 27, 70, 150\\
            Magic & 45, 61, 87 & 44, 57, 82 & 45, 58, 88 & 46, 59, 86 & 45, 60, 86\\
            ATACH2 & 84, 100, 120 & 82, 99, 122 & 81, 97, 125& 79, 98, 124 & 82, 103, 128\\
            ERICH & 70, 87, 110 & 66, 82, 111 & 51, 82, 104 & 62, 80, 108 & 62, 80, 117\\
            \bottomrule
        \end{tabular}
        \end{center}
        \caption{Number $n$ of examples and DCR to \textbf{held out dataset}. 25\%, 50\% (Median), 75\% quantile.}
        \label{table:num-shots-dcr2}
\end{table*}

\section{Cost Analysis}
The section demonstrates some examples of the time cost of our framework on the real world datasets. We provide both the training time and testing time in Table \ref{tab:cost}.
\begin{table*}[t]
    \centering
    \begin{tabular}{cccccc}
    \toprule
    Dataset & Sample Size & Number of Epochs & Number of Examples Per Call & Training Time & Inference Time\\
    \midrule
    Asia &  50 & 10 & 4 & 13.5 & 5.9\\
    Asia &  200 & 4 & 4 & 24.8 & 3.8\\
    ERICH& 100 & 5 & 1 & 62.7 & 14.6\\
    ERICH& 200 & 4 & 1 & 92.5 & 23.7\\
    \bottomrule
    \end{tabular}
    \caption{Examples of computational cost analysis. (Unit: Minutes)}
\label{tab:cost}
\end{table*}

\label{sec:appendix}

This is an appendix.

\end{document}